%% file: ijcai19.tex
\documentclass{article}
\usepackage{ijcai19}

\usepackage{times}
\usepackage{soul}
\usepackage{url}
\usepackage[hidelinks]{hyperref}
\usepackage[utf8]{inputenc}
\usepackage[small]{caption}
\usepackage{graphicx}
\usepackage{amsmath}
\usepackage{booktabs}
\usepackage{algorithm}
\usepackage{algorithmic}
\input{head.tex}

\title{A Deep, Information-theoretic Framework  for Robust Biometric Recognition}

\iftrue
\author{
Renjie Xie$^{1,3}$
\and
Yanzhi Chen$^{2,3}$\and
Yan Wo$^{3,}$\footnote{corresponding author}\and
Qiao Wang$^{1}$
\affiliations
$^1$Southeast University, Nanjing, China\\
$^2$The University of Edinburgh, Edinburgh, UK\\
$^3$South China University of Technology, Guangzhou, China\\
\emails
\text{renjie\_xie}@seu.edu.cn, \quad
s1788325@inf.ed.ac.uk, \quad
woyan@scut.edu.cn, \quad
qiaowang@seu.edu.cn
}
\fi

\begin{document}

\maketitle

\input{sections/abstract.tex}
\input{sections/1-introduction.tex}

\input{sections/2-attack.tex}

\input{sections/3-method.tex}

\input{sections/4-experiments.tex}

\input{sections/5-conclusion.tex}
\clearpage

\appendix

\bibliographystyle{named}
\bibliography{ijcai19}

\end{document}


\maketitle

\vspace{-2.0cm}

\appendix

\section{Derivation of the lower bound of the learning objective}
We here provide the details for deriving equation (12), the lower bound of our learning objective $\mathcal{L'}$. The derivation is similar to that of the original DVIB literature. Remark that the objective is:
\begin{equation}
    \text{maximize} \quad \mathcal{L'} =  I(Z';Y)-\beta \cdot I(Z;X)
    \label{formula:IB}
\end{equation}
Here, as in DVIB, we make the assumption that the joint distribution $p(\vecx, y, \vecz)$ is factorized as
\begin{equation}
    p(\vecx,y,\vecz) = p(\vecx)p(y|\vecx)p(\vecz|\vecx)
\end{equation}
which means that the corresponding directed graph is $Z \leftarrow X \rightarrow Y$. 

The lower bound for the first term $I(Z';Y)$ is:
\begin{equation}
\begin{split}
    I(Z'; Y) &= \iint p(y, \vecz') \log \frac{p(y, \vecz')}{p(y)p(\vecz)} dy d\vecz' \\
    &= \iint p(y, \vecz') \log p(y|\vecz')dy d\vecz - H[Y] \\
    &= \int p(\vecz') \Big[ \int p(y|\vecz') \log p(y|\vecz') dy \Big] d\vecz' - H[Y] \\
    &\geq \int p(\vecz') \Big[ \int p(y|\vecz') \log q(y|\vecz') dy \Big] d\vecz' - H[Y] \\
    &= \iint p(y, \vecz') \log q(y|\vecz') dy d\vecz' - H[Y] \\
    &= \iiint p(\vecx, y) p(\vecz'|\vecx, y) \log q(y|\vecz') dy d\vecz' d\vecx  - H[Y] \\
    &= \iint p(\vecx, y) \Big[\int p(\vecz'|\vecx) \log q(y|\vecz')\Big d\vecz'  \Big] d\vecx dy - H[Y] \\
    &\approx \frac{1}{n} \sum^n_{i=1}  \mathbb{E}_{p(\vecz'|\vecx_i)} \Big[ \log q(y_i|\vecz')  \Big] - H[Y].
\end{split}
\end{equation}
The inequality is due to $\text{KL}[p(y|\vecz') ||q(y|\vecz')] \geq 0$. 

The upper bound for the second term $I(X;Y)$ is:
\begin{equation}
\begin{split}
   I(X; Z) &=  \iint p(\vecx, \vecz) \log \frac{p(\vecx, \vecz)}{p(\vecz)p(\vecx)} d\vecx d\vecz \\
    &= \iint p(\vecz,\vecx) \log p(\vecz|\vecx)d\vecx d\vecz - \int p(\vecz)\log p(\vecz) d\vecz \\
    &\leq \iint p(\vecz, \vecx) \log p(\vecz|\vecx)d\vecx d\vecz - \int p(\vecz) \log q(\vecz) d\ \vecz\\
    &= \iint p(\vecx) p(\vecz|\vecx) \log \frac{p(\vecz|\vecx)}{q(\vecz)} d\vecx d\vecz \\
    &\approx \frac{1}{n} \sum^n_{i=1} \Big[\text{KL}[p(\vecz|\vecx_i) || q(\vecz) \Big].
\end{split}
\end{equation}
The inequality is due to $\text{KL}[p(\vecz)||q(\vecz)] \geq 0$. Putting all together yields
\begin{equation}
    \mathcal{L}' \geq \frac{1}{n} \sum^n_{i=1} \Big[ \mathbb{E}_{p(\vecz'|\vecx_i)} \Big[ \log q(y_i|\vecz') \Big]  - \beta \cdot \text{KL}[p(\vecz|\vecx_i)||q(\vecz)]  \Big] - H[Y]
\end{equation}
and since $H[Y]$ is a constant, we are safe to drop it from the objective for optimization.

\section{Details of the CNNs}
The convolutional neural networks (CNN) employed in the experiments contain 20 layers that are grouped into 5 stages, as summarized in Figure 1. 

\begin{figure*}[h]
    \centering
        \includegraphics[width=1.0\textwidth]{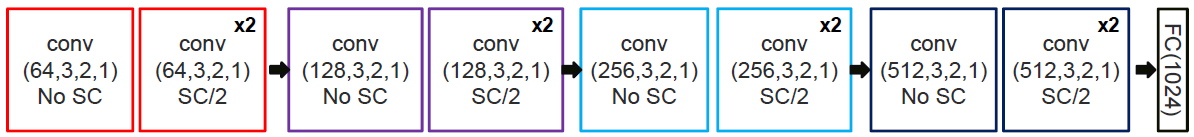}
        \caption{The detailed architecture of the CNNs employed in the experiments. }
\end{figure*}
in which: 
\begin{itemize}[leftmargin=*]
    \item \emph{Conv} means the convolutional layer, the figures $(n,s,p,r)$ mean that there are $n$ filters with $s \times s$ size in the layer, the stride is $p$, and the padding s $r$; 
    \item \emph{No SC} means that there is no short cut connection and \emph{SC/2} means that there is a short cut connection between every two layers; 
    \item \emph{FC} indicates the fully connected layer. There are 1024 units in the FC layer. 
    \item Parametric Rectified Linear Unit (pReLU) is adopted as the non-linearity in the network. The activation function of pReLU is:
    \begin{equation}
      pReLU(x) = 
      \left\{
      {
      \begin{array}{cl}
        x & \quad \text{if  } x>0  \\
       ax  & \quad \text{if  } x\leq0 
      \end{array}
      } \right. 
    \end{equation}
    where $a$ is a learnable parameter. The initial value of $a$ is set to be $a = 0.25$.  
\end{itemize}
The weights of the CNN would be jointly trained with that in the subsequent network through BP. 

\quad \\

\section{Details of the modified Carlini-Wanger algorithm}
Here we provide the details of the modified Carlini-Wanger attack for constructing adversarial biometrics in our experiment. Remark that to find the adversarial biometric $\tilde{\vecx}_1$ we need to optimize the following objective:
\begin{equation}
    J'_{\text{adv}}(\tilde{\vecx}_1) = \| \vecx_1 - \tilde{\vecx}_1 \|_2 + \lambda \cdot \cos(f(\tilde{\vecx}_1), f(\vecx_2))
    \label{formula:carlini-attack}
\end{equation}
which is subject to the constraint $x_k \in [0,1]$. To remove this constraint we reparameterize each $\vecx$ as
\begin{equation}
    \vecx = h(\vecv) = \frac{1}{2} \tanh{\vecv} + 1
\end{equation}
with which we can rewrite (7) as:
\begin{equation}
    J'_{\text{adv}}(\tilde{\vecz}_1)) = \| h(\vecz_1) - h(\tilde{\vecz}_1) \|_2 + \lambda \cdot \cos(f(h(\tilde{\vecz}_1)), f(h(\vecz_2)))
    \label{formula:carlini-attack}
\end{equation}
and we can now learn $\tilde{\vecz_1}$ by gradient descent.

For the selection of $\lambda$, we find the optimal value of $\lambda$ by an iterative procedure. Starting from $\lambda = 1$, we will update the value of $\lambda$ as follows:
\begin{equation}
      \lambda = 
      \left\{
      { 
      \begin{array}{cl}
        10\lambda & \quad \text{if the solved } \tilde{\vecz}_1 \text{ in (9) satisfies: } \cos(f(h(\tilde{\vecz}_1)), f(h(\vecz_2))) \leq T  \\
        \lambda/2 & \quad \text{if the solved } \tilde{\vecz}_1 \text{ in (9) satisfies: } \cos(f(h(\tilde{\vecz}_1)), f(h(\vecz_2))) \geq T
      \end{array}
      } \right. 
\end{equation}
This procedure is repeated until converge. $T$ is selected as the threshold at which the equal error rate (EER) is attained. All optimization is done by Adam with its default settings.


%% file: head.tex
\usepackage[utf8]{inputenc}
\usepackage{amsmath,amssymb,amsthm}

\usepackage{float}
\usepackage{subfig}
\usepackage{dsfont}
\usepackage{multirow}
\usepackage{booktabs}
\usepackage{graphicx}
\usepackage{enumitem}
\usepackage{multicol}
\usepackage{mathrsfs}
\usepackage{bbm}
\usepackage{setspace}
\usepackage{amsfonts}
\usepackage{tikz}
\usetikzlibrary{bayesnet}
\usepackage[export]{adjustbox}

\DeclareMathOperator*{\argmin}{arg\,min}

\newcommand{\vecx}{\mathbf{x}} 

\newcommand{\vecz}{\mathbf{z}}
\newcommand{\vecc}{\mathbf{c}}
\newcommand{\vecu}{\mathbf{u}}
\newcommand{\vecv}{\mathbf{v}}
\newcommand{\W}{\mathbf{W}}

\newcommand{\X}{\mathbf{X}}

\theoremstyle{definition}

%% file: sections/abstract.tex
\begin{abstract}
\noindent Deep neural networks (DNN) have been a \emph{de facto} standard for nowadays biometric recognition solutions. A serious, but still overlooked problem in these DNN-based recognition systems is their vulnerability against adversarial attacks. Adversarial attacks can easily cause the output of a DNN system to greatly distort with only tiny changes in its input. Such distortions can potentially lead to an unexpected match between a valid biometric and a synthetic one constructed by a strategic attacker, raising security issue. In this work, we show how this issue can be resolved by learning robust biometric features through a deep, information-theoretic framework, which builds upon the recent deep variational information bottleneck
method but is carefully adapted to biometric recognition tasks. Empirical evaluation demonstrates that our method not only offers stronger robustness against adversarial attacks but also provides better recognition performance over state-of-the-art approaches.
\end{abstract}

%% file: sections/1-introduction.tex
\section{Introduction}
Let us consider the following scenario in real-world biometric recognition. In particular, we will be focusing on the case of \emph{deep biometric recognition}, where deep neural network (DNN) is employed as flexible feature extractors in the database. Recognition is then done by comparing the distance $D$ between two biometric $\vecx_1, \vecx_2$ in the feature space:
\begin{equation}
        \begin{cases}
          y_1 = y_2 & \quad \text{if} \quad D(f(\vecx_1), f(\vecx_2)) \le T \\
          y_1 \neq y_2 & \quad \text{if} \quad D(f(\vecx_1), f(\vecx_2)) > T 
        \end{cases}
        \label{formula:biometric-recognition}
\end{equation}  
where $y$ denotes the identity (class) of the biometric, $T$ is the decision threshold, $f: \mathbb{R}^{K} \to \mathbb{R}^{k}$ is the DNN used to extract features and $D(\cdot, \cdot)$ is some distance metric in the feature space (e.g Euclidean distance or cosine distance). Therefore two biometrics are considered to be matched with each other if the distance between their features falls within $T$. Now assume that a malicious attacker, who is also a \emph{valid} user in the system, is going to `masquerade' some other person in the same system. The question here is: can the attacker modify his own biometric image $\vecx$ with only little efforts $\Delta \vecx$, so that the system will mismatch him as some other user in the same system? If the answer is yes, the attacker can then be granted full authority belonged to that user ------ a security disaster.

The above scenario is indeed not highly hypothetical due to the advent of \emph{adversarial attacks} in recent machine learning. Adversarial attacks are algorithms designed to strike the stability of neural networks. Such algorithms can cause the output of a neural network to greatly distort with only tiny variations in its input, leading to sub-optimal or even completely wrong decisions \cite{szegedy_intriguing_2013}. For example, under adversarial attacks, a well-trained deep neural network can be fooled by a modified cookie picture and ridiculously classifies it as a dog face. The source of adversarial attacks is still an open problem \cite{szegedy_intriguing_2013,goodfellow_explaining_2014,moosavi_deepfool_2016}, but one thing is clear to us: it alerts us to blindly adopting DNN in those security-sensitive applications. Back to biometric recognition, if the output distortion caused by adversarial attacks are so large that it incurs a unexpected match between the attacker's biometric and that of other users, security issue raises. Therefore, while offering human-surpassing recognition performance, the robustness of existing DNN-based recognition systems \cite{sun_deep_2014,amos_openface_2016,sun_deepid3_2015} are indeed susceptible and may need careful re-examination.

Very few efforts have been devoted to study the effect of adversarial attacks in biometric recognition. In this work, we investigate how adversarial attacks can raise serious security problems in existing deep biometric recognition systems as well as exploring its solution. We begin by extending existing adversarial attacking algorithm from label space to feature space and then use it to attack DNN-based biometric recognition system. We subsequently show that how this kind of attacks can be effectively defended by a new deep feature extraction framework, which is evolved from the recent deep variational information bottleneck method \cite{alemi_deep_2017} but is augmented with two improvements: a novel sphere projection operation and an adversarial learning strategy. The resultant method, which is coined as deep variational sphere projection (DVSP), provides both strong robustness to adversarial attacks as well as satisfactory recognition performance, hence is more preferable in real-world applications, as we confirm by extensive experiments.

The rest of the paper are organized as follows. Section 2 formulates the problem as well as briefly introduces the adversarial algorithm used to attack DNN-based biometric recognition system. Section 3 elaborates the details of the proposed DVSP framework. Section 4 conducts a set of experiments on both the toy MNIST dataset and the real-world LFW dataset. We finally conclude the paper in Section 5.

%% file: sections/2-attack.tex
\section{Problem formulation}

Below, we first briefly revisit the background knowledge of adversarial attacks in recent machine learning research, then illustrate how to extend these attacks to the context of biometric recognition, which triggers the motivation of this work.

\subsection{Adversarial attacks in label space}

\newcommand{\Din}{d}
\newcommand{\Dout}{D} 

Generally speaking, given a classifier $f(\cdot)$ and two input images $\vecx_1$, $\vecx_2$ from different classes (e.g dogs v.s cats), adversarial attacks seek $\vecx_1$'s minimal modified version $\tilde{\vecx}_1$ such that it can cause $f(\cdot)$ to mis-classify $\tilde{\vecx}_1$ as $\vecx_2$. It can be formulated as the following constrained optimization problem:
\begin{equation}
        \begin{split}
            \tilde{\vecx}_1=& \mathop{\argmin}_{\vecx}{\Din(\vecx, \vecx_1)}\\
            s.t. \quad& f(\tilde{\vecx}_1) = f(\vecx_2) \\
            \qquad& y_1 \neq y_2
        \end{split}
        \label{formula:adv_attack}
\end{equation}
where $\Din(\cdot, \cdot)$ are some metric that measures the changes in the classifier's input space. We say that $f(\cdot)$ is vulnerable under adversarial attacks if $d(\tilde{\vecx}_1, \vecx_1)$ for the found $\tilde{\vecx}$ is small. Diverse attacking algorithms have been developed under different metric $\Din$; see e.g literature \cite{yuan2019adversarial} for a recent review. In this work, we mainly focus on the case $\Din(\tilde{\vecx}, \vecx) = \| \tilde{\vecx} - \vecx \|_2$ due to its popularity and simplicity.

\subsection{Adversarial attacks in feature space}
Existing works in adversarial attacks mainly consider the \emph{classification} scenario where $f(\vecx)$ outputs the class label i.e it is defined in the label space. However it is indeed straightforward to extend them to the feature space. To see this, we first relax \eqref{formula:adv_attack} as:
\begin{equation}
        \begin{split}
            \tilde{\vecx}_1=& \mathop{\argmin}_{\vecx}{\|\vecx -  \vecx_1\|_2}\\
            s.t. \quad&\Dout(f(\tilde{\vecx}_1), f(\vecx_2)) \le T\\
            \qquad& y_1 \neq y_2
        \end{split}
        \label{formula:adv_attack2}
\end{equation}
where $\Dout(\cdot, \cdot)$ is a distance metric in the output space and $T$ is a threshold. One can see that if we broaden the family of $f(\cdot)$ from classifier to feature extractor and pick appropriate distance metric $\Dout(\cdot, \cdot)$ we can immediately adapt existing adversarial attacks to feature space. 

Taking Carlini-Wanger attack \cite{carlini_towards_2017}, one of the most powerful adversarial attacking algorithm as example, it originally finds $\tilde{\vecx}_1$ by optimizing a regularized optimization problem (assuming that $\Din(\tilde{\vecx}, \vecx) = \| \tilde{\vecx} - \vecx \|_2$):
\[
    J_{\text{adv}}(\tilde{\vecx}_1) = \| \vecx_1 - \tilde{\vecx}_1 \|_2 + \lambda \cdot \max_i | s(\vecx_1)_i - s(\tilde{\vecx}_1)_{y_2}|
    \label{formula:carlini-attack}
\]
where $\lambda$ is the Lagrangian multiplier and $s(\vecx)$ is the softmax scores of the classifier $f(\vecx)$. When adapting it to feature space, we can replace the regularized term by e.g the cosine distance between the extracted features: 
\[
    J'_{\text{adv}}(\tilde{\vecx}_1) = \| \vecx_1 - \tilde{\vecx}_1 \|_2 + \lambda \cdot \cos(f(\tilde{\vecx}_1), f(\vecx_2))
    \label{formula:carlini-attack}
\]
with which we can solve $\tilde{\vecx}_1$ by the same optimization procedure as before. Many other attacking algorithms in label space can be adapted to feature space in a similar way.

\begin{figure}[H]
                \centering
                \includegraphics[width=0.85\linewidth]{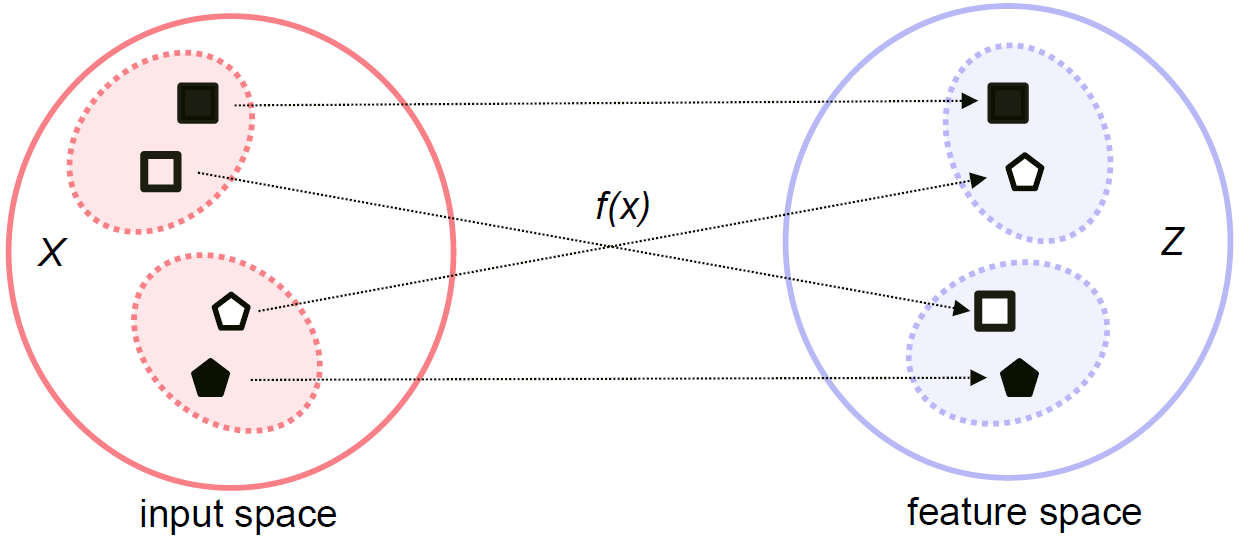}
                \caption{A conceptual demonstration on how adversarial attacks can be exploited to attack DNN-based biometric recognition system. Solid items: valid biometric in the database. Empty items: fake biometric constructed by adversarial attacks from the valid biometric. The dashed ellipsoids represent small $T$-balls in the two spaces. While the difference between the original biometric and the adversarial biometric is small in the original space, they are sufficient to cause a mismatch in the feature space.}
                \label{fig:demonstration}
\end{figure}

The existence of adversarial attacks in the feature space clearly raises a prominent security problem for DNN-based recognition systems: a strategic attacker may make use of adversarial attacks to construct fake biometric from existing ones whose features are close enough to that of other users (Figure \ref{fig:demonstration}). If this is true, the attacker can be mis-recognized as other users and be granted full authority belonging to others. It is then natural to ask how to build a feature extractor $f(\cdot)$ that are robust against adversarial attacks. Before answering this question, we need a principle way to measure the robustness of $f(\cdot)$ in the feature space, as existing robustness evaluation metric \cite{huang_safety_2017,bastani_measuring_2016,carlini_towards_2017} only work for the label space. Here, we suggest to quantify the robustness of $f(\cdot)$ by the `hardness' of finding $\tilde{\vecx}_1$  in \eqref{formula:adv_attack2}:
\begin{equation}
            \begin{split}
                \mathcal{H}_{1,2}(f) = \frac{1}{n_1 n_2} \sum_{\vecx_1 \in \X_1} \sum_{\vecx_2 \in \X_2} \inf_{D(f(\tilde{\vecx}_1), f(\vecx_2)) \le T}
                \left\|\tilde{\vecx}_1 - \vecx_1\right\|_2
            \end{split}
            \label{formula:pairwise-robustness-index}
\end{equation}
where $\X_1$ and $\X_2$ are the collection of biometric images belonging to individuals 1 and 2, $n_1, n_2$ are the cardinal of $\mathbf{X}_1$ and $\mathbf{X}_2$, respectively. Therefore we measure the robustness of $f(\cdot)$ as the minimal effort required to make the system mis-recognize $\vecx_1$ as $\vecx_2$ and average it across all $(\vecx_1, \vecx_2)$ pairs. Note that $\mathcal{H}_{1,2}(f)$ is defined for the binary-class case but it is trivial to generalize it to multi-class case.  

Our goal here is hence to design a feature extractor $f(\cdot)$ with high robustness index \eqref{formula:pairwise-robustness-index} but still guarantees good recognition performance. Next, we show that how this goal can be achieved by learning robust biometric features through a deep, information-theoretic framework, which is evolved from the recent deep variational information bottleneck (DVIB) framework but are augmented with two important modifications designed for biometric recognition tasks.

%% file: sections/3-method.tex
\begin{figure*}[!t]
            \hspace{0.01\linewidth}
            \subfloat[Features in DVIB]{
                    \centering
                    \label{fig:vsp-a}
                    \begin{minipage}[t]{0.23\linewidth}
                    \centering
                     \includegraphics[height=0.90\linewidth]{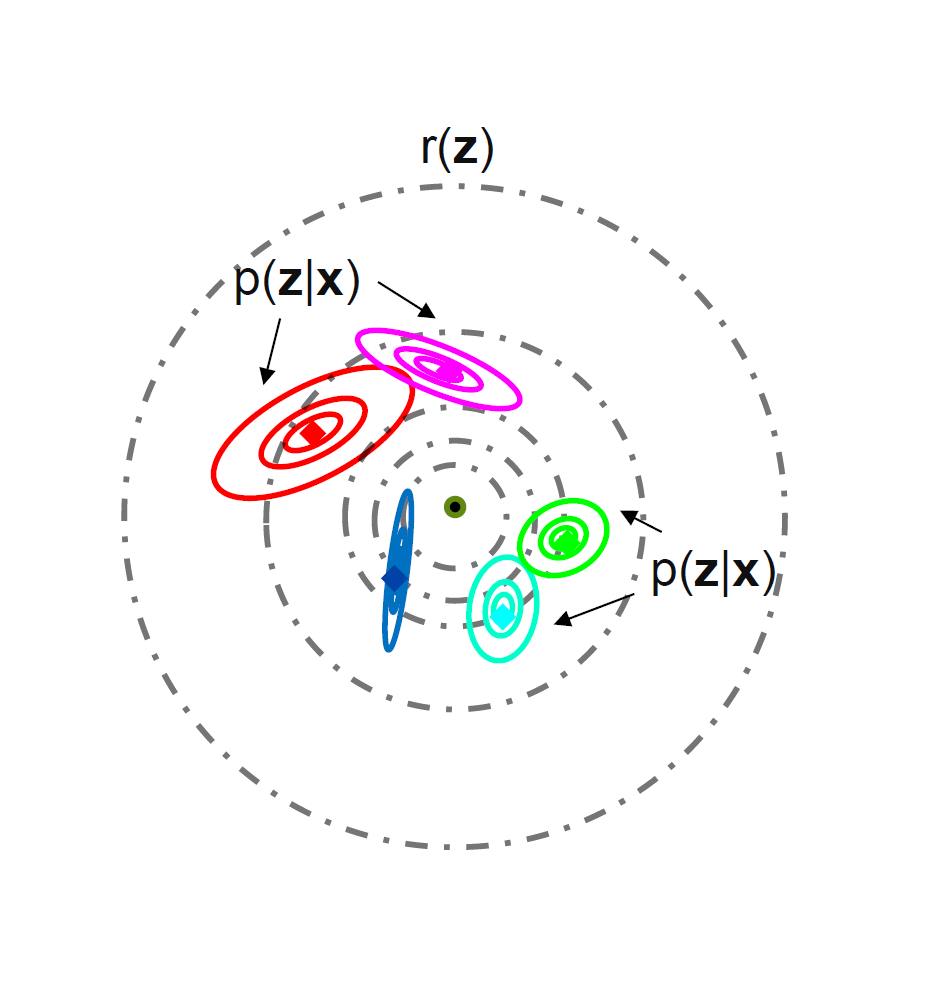}
                    \end{minipage}
            }
            \subfloat[Sphere projection]{
                    \centering
                    \label{fig:vsp-b}
                    \begin{minipage}[t]{0.23\linewidth}
                    \centering
                    \includegraphics[height=0.90\linewidth]{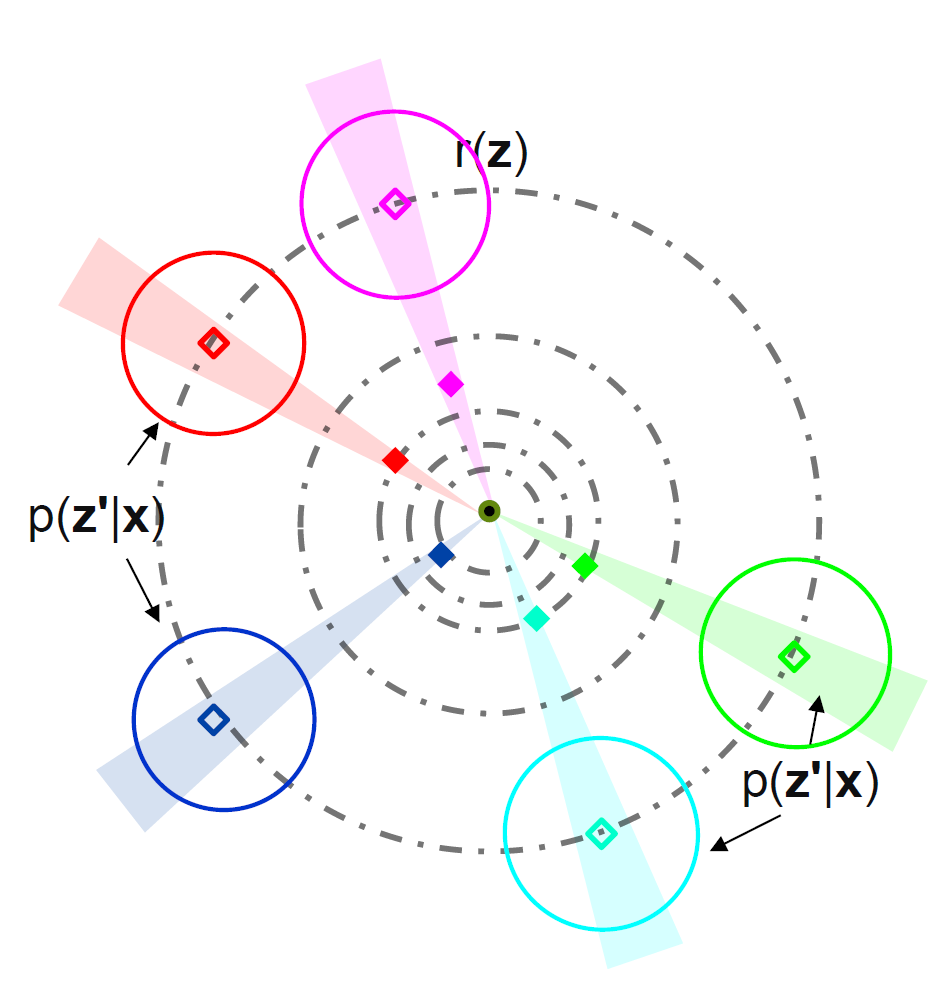}
                    \end{minipage}
            }
            \subfloat[Adversarial learning]{
                    \centering
                    \label{fig:vsp-c}
                    \begin{minipage}[t]{0.23\linewidth}
                    \centering
                     \includegraphics[height=0.90\linewidth]{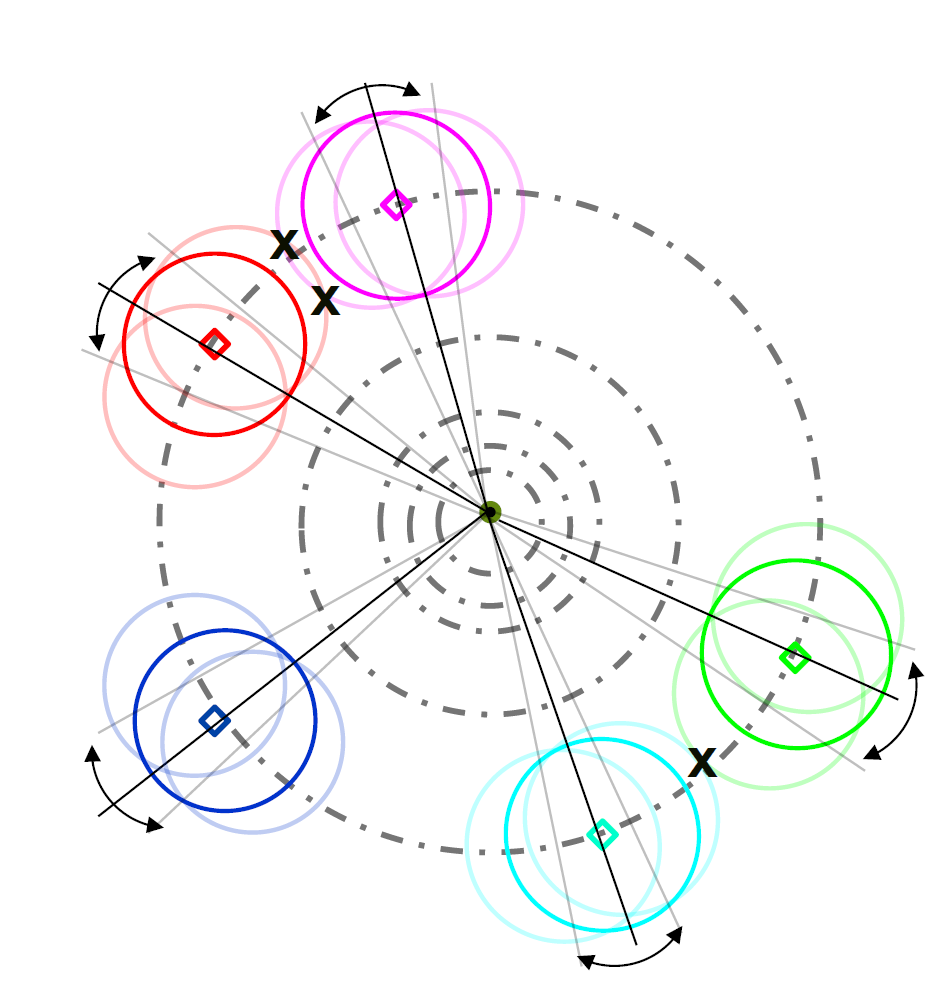}
                    \end{minipage}
            }
            \hspace{0.00\linewidth}
            \subfloat[Resultant features]{
                    \centering
                    \label{fig:vsp-d}
                    \begin{minipage}[t]{0.23\linewidth}
                    \centering
                    \includegraphics[height=0.90\linewidth]{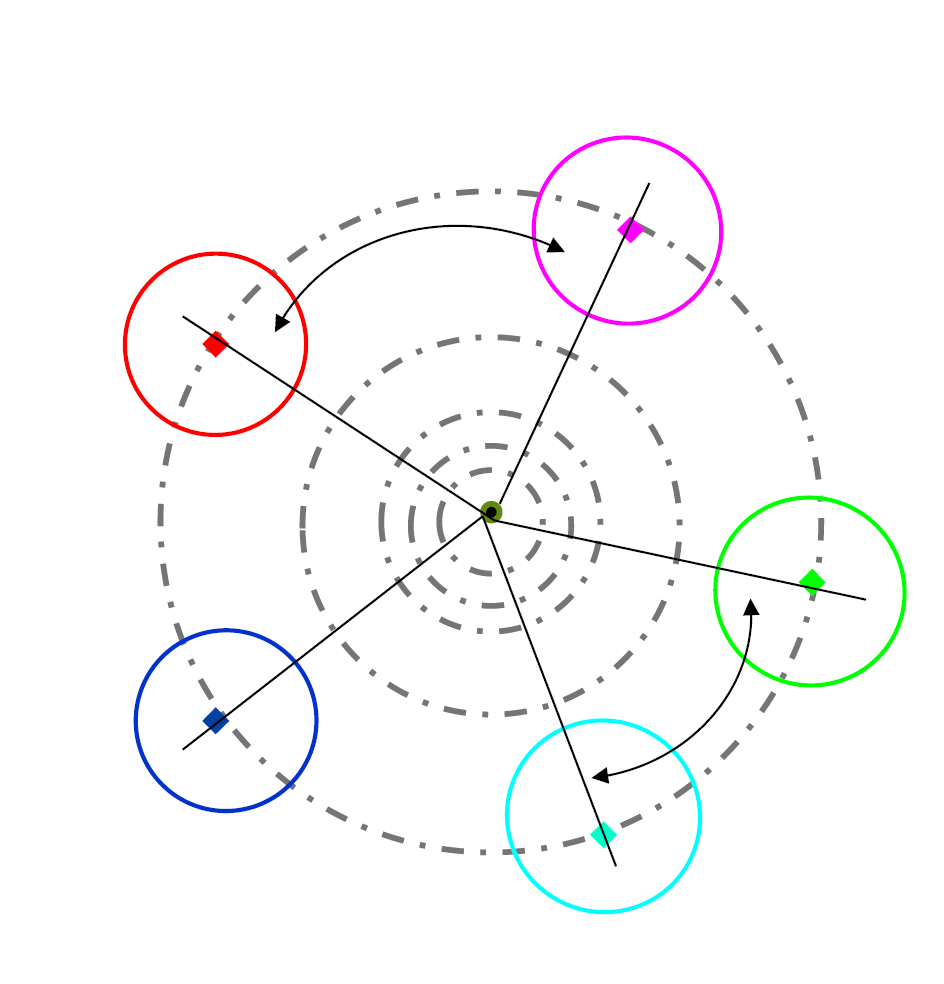}
                    \end{minipage}
            }
            \caption{Demonstrating the feature extraction process of DVSP. Black dashed circles: the contour plot of $r(\vecz)$. Colored ellipses: contour plot of $p(\vecz|\vecx)$. Colored circles: contour of $p(\vecz'|\vecx)$. Solid diamonds: the features in DVIB. Empty diamonds: the projected features in DVSP. } 
            \label{fig:3_2}
\end{figure*}

\section{Method}
From an information-theoretic perspective, if the extracted features, $Z$, forget as much irrelevant information as possible in the input biometric $X$, then it shall be robust to small permutations in the inputs. On the other hand, for good recognition performance we hope that the learned features $Z$ are still informative about $Y$, the identity of the biometric owner. Together, we can formulate an information-bottleneck (IB) principle \cite{tishby2000information} for feature extraction:
\begin{equation}
    \text{maximize} \quad \mathcal{L} =  I(Z;Y)-\beta \cdot I(Z;X)
    \label{formula:IB}
\end{equation}
Unfortunately, exact optimization of \eqref{formula:IB} is difficult due to the intractable integrals in mutual information computation. Below, we first show how to optimize \eqref{formula:IB} approximately by the recent deep variational information bottleneck framework (DVIB) \cite{alemi_deep_2017}, then adapt it to biometric recognition context.

\subsection{Deep variational information bottleneck}
Rather than optimizing (\ref{formula:IB}) directly, DVIB suggests to instead optimize its variational lower bound :
\begin{equation}
        \begin{split}
        \mathcal{L} &\geq 
        \frac{1}{n}\sum\limits_{i=1}^n \mathbb{E}_{p(\vecz|\vecx_i)} \Big[\log q(y_i|\vecz) \Big]
        - \beta \text{KL}\Big[p(\vecz|\vecx_i) \| r(\vecz) \Big]
        \end{split}
      \label{formula:IB2}
\end{equation}
where $q(y|\vecz)$ and $q(\vecz)$ are the variational approximations to the distributions $p(y|\vecz)$ and $p(\vecz)$ respectively. See \cite{alemi_deep_2017} for the derivation. In DVIB, these distributions are modeled as:
\begin{equation}
\begin{split}
        p(\vecz|\vecx) = \mathcal{N}(\vecz; \mu(\vecx), \Lambda(&\vecx)), \qquad q(\vecz) = \mathcal{N}(\vecz; \mathbf{0}, \mathbf{I}) \\
        q(y=j|\vecz) &= \frac{e^{g_j(\vecz)}}{\sum^K_{k=1} e^{g_k(\vecz)}}
        \label{formula:q,p,r}
\end{split}
\end{equation}
where the mean $\mu(\vecx)$ and the covariance $\Lambda(\vecx)$ of $p(\vecz|\vecx)$ are computed by an \emph{encoder} network $f$ with weights $\W_f$:
\begin{equation}
          f(\vecx; \mathbf{W}_f) = \{\mu(\vecx), \Lambda(\vecx)\}
\end{equation}
whereas the softmax score $g_k(\vecz)$ in $q(y=j|\vecz)$ is given by another \emph{decoder} network $g$ with weights $\W_g$:
\begin{equation}
        g(\vecz; \mathbf{W}_g) = \{g_{k}(\vecz) \}^K_{k=1}
\end{equation}
The weights in the two networks $f$ and $g$ are then jointly learned by optimizing \eqref{formula:IB2}. After training, we take $\mu(\vecx)$ in the encoder network as the features of $\vecx$ i.e $\vecz = \mu(\vecx)$.

\subsection{Deep variational sphere projection}
DVIB allows us to learn $Z$ in \eqref{formula:IB} efficiently, however it is originally designed for classification tasks rather than biometric recognition. Mainstream biometric recognition solutions usually learn biometric features with a large inter-class margin explicitly so as to achieve good recognition performance in \eqref{formula:biometric-recognition} (see e.g \cite{chopra_learning_2005}, \cite{schroff_facenet_2015}, \cite{wen_discriminative_2016} and \cite{liu2016large}). In DVIB, however, $\mu(\vecx)$ of each classes has to stay near the origin due to the KL term in \eqref{formula:IB2}, making their Euclidean distance too close to each other (Figure \ref{fig:3_2}.a) and hence small inter-class margin. We introduce two important modifications to DVIB to resolve this problem. \\

\begin{figure*}[t]
    \centering
        \includegraphics[width=0.81\textwidth]{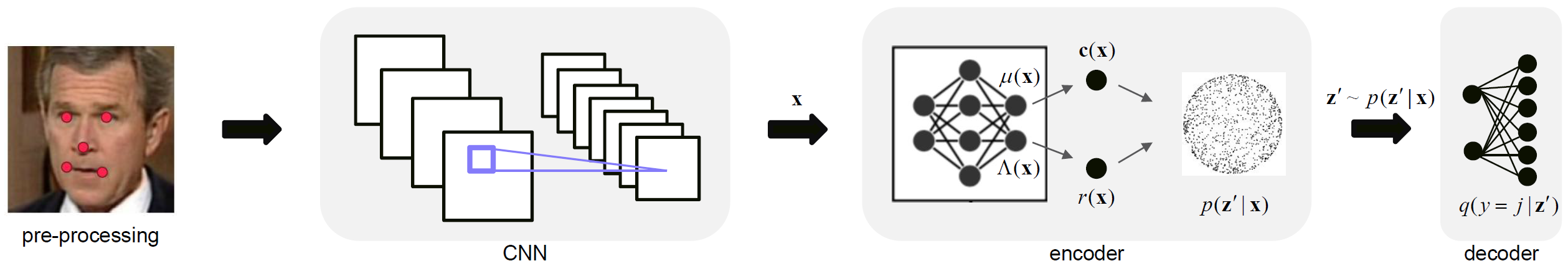}
        \caption{An overview of the proposed DVSP framework for  robust biometric features extraction. }
        \label{figure:nn-architecture}
\end{figure*}

\noindent \textbf{Sphere projection}. The first modification  is a novel projection operation. The idea is to `project' the features in DVIB to the surface of a hypersphere, where the projected features are allowed to have large \emph{angular} margin regardless of the KL term (Figure \ref{fig:3_2}.b). The projection operation is defined as:
\begin{equation}
    \vecz' = R \cdot\frac{ \vecz} {\|\vecz\|}
    \label{formula:z'}
\end{equation}
Therefore the new features $\vecz'$ are on the surface of a hypersphere whose center is at the origin and has radius $R$. Similar to \eqref{formula:IB}, we can set up an IB principle for learning $Z'$:
\begin{equation}
        \begin{split}
         \text{maximize} \quad \mathcal{L}' &=  I(Z';Y)-\beta \cdot I(Z';X) 
        \end{split}
        \label{formula:IB3}
\end{equation}
whose lower bound is:
\begin{equation}
        \begin{split}
        \mathcal{L}' &\geq I(Z';Y)- \beta \cdot I(Z;X) \\
        &\geq \frac{1}{n} \sum \limits_{i=1}^n \underbrace{\mathbb{E}_{p(\vecz'|\vecx_i)} \Big[\log q(y_i|\vecz') \Big]}_{\text{likelihood term}}
        - \beta \underbrace{\text{KL}\Big[p(\vecz|\vecx_i) \| q(\vecz) \Big]}_{\text{KL term}} 
        \end{split}
      \label{formula:IB4}
\end{equation}
The first inequality above is given by the data processing inequality and the second inequality is given by the variational lower bound. Learning now amounts to optimizing \eqref{formula:IB4} instead of \eqref{formula:IB2}. Here, $p(\vecz'|\vecx)$ is modeled as a special `shell distribution' whose probability mass is uniformly spread on the surface of a hypersphere:
\begin{equation}
      p(\vecz'|\vecx) 
      \left\{
      {
      \begin{array}{cl}
        \propto 1,  & \quad \text{if  } \|\vecz' - \vecc(\vecx) \| = r(\vecx) \\
       = 0  & \quad \text{otherwise}
      \end{array}
      } \right. 
    \label{formula:p(z'|x)}
\end{equation}
and the center $\vecc(\vecx)$ and radius $r(\vecx)$ of the hypersphere are:
\begin{equation}
    \hspace{0.10\linewidth}  \vecc(\vecx) = \mu(\vecx)/\|\mu(\vecx)\|, \quad r(\vecx) = \frac{1}{k} \sum_{i=1}^k \Lambda_{i,i}(\vecx)
    \label{formula:sphere}
\end{equation}

The reason that why sphere projection helps in generating large-margin features is as follows. To maximize \eqref{formula:IB4}, the optimizer pursuits a large value for the expectation term $\mathbb{E}_{p(\vecz'|\vecx_i)} [\log q(y_i|\vecz')]$. As each $\vecz' \sim p(\vecz'|\vecx_i)$ in the term is randomly sampled from the surface of the hypersphere, if the centers of inter-class hyperspheres are too close to each other, those $\vecz'$ sampled from different classes would easily crash together, leading to low $\log q(y|\vecz')$ values. As a consequence, the optimizer would drive the surface of these hyperspheres to be far way from each other and ultimately, resulting in a large angular margin between their centers (Figure 2.b). 

In practice, we can compute the gradient of \eqref{formula:IB4} as follows. Denote $\W = \{\W_f, \W_g\}$ as the parameters to be optimized:

\newcommand{\spheredist}{\mathcal{U}_{\text{sphere}}( \mathbf{0}, 1)}

\begin{itemize}[leftmargin=*]
\item \emph{Likelihood term}. We compute it by the reparameterization trick \cite{kingma_auto-encoding_2013}. Let $\vecu \sim \spheredist$ be the sample randomly drawn from the surface of an unit sphere. Each sample $\vecz' \sim p(\vecz'|\vecx_i)$ can be rewritten as:
\[
    \vecz' \sim p(\vecz'|\vecx_i) \quad \Leftrightarrow \quad \vecz' =  r(\vecx_i)\cdot\vecu + \vecc(\vecx_i)
\]
where $r$ and $\vecc$ are defined in \eqref{formula:sphere}.  We can then rewritten the gradient of the expectation term as:
\begin{eqnarray*}
        \lefteqn{ \nabla_{\mathbf{W}} \mathbb{E}_{\vecz' \sim p(\vecz'|\vecx_i)}\Big[\log q(y_i|\vecz') \Big]  } \\
        & & = \mathbb{E}_{\vecu \sim \spheredist}\Big[\nabla_{\mathbf{W}}\log q(y_i|r(\vecx_i)\vecu + \vecc(\vecx_i) ) \Big]  \\
        & & \approx \frac{1}{m} \sum_{l=1}^m \nabla_{\mathbf{W}}\log q(y_i|r_i\vecu_l + \vecc_i ), \enspace \vecu_l \sim \spheredist
\end{eqnarray*}
where, to obtain $\vecu \sim \spheredist$, one can simply first sample $\vecv \sim \mathcal{N}(\mathbf{0}, \mathbf{I})$, then get $\vecu$ by $\vecu = \vecv/\| \vecv \|$. 


\item \emph{KL term}. This term can be computed easily since both $p(\vecz|\vecx_i)$ and $r(\vecz)$ are Gaussian. Here, as in DVIB, we assume that the matrix $\Lambda(\vecx_i)$ in $p(\vecz|\vecx_i)$ is diagonal, yielding
\[
\begin{split}
\nabla&_{\W}  \text{KL} \Big[p(\vecz|\vecx_i) \| q(\vecz) \Big]  \\
        & = \frac{1}{2} \nabla_{\W} \Big[ \log \det(\Lambda(\vecx_i)) + \text{Tr}({\Lambda(\vecx_i))}  +  \|\mu(\vecx_i)\|_2^2 \Big] \\
        & = \frac{1}{2} \nabla_{\W} \Big[ \sum_l^k \log \Lambda_{l,l}(\vecx_i) + \sum_l^k \Lambda_{l,l}(\vecx_i) + \| \mu(\vecx_i)\|_2^2 \Big] 
\end{split}
\]
\end{itemize}
Unifying results from both terms yields the gradient $\nabla_{\W} \mathcal{L}'$.

\noindent \textbf{Adversarial learning}. The second modification in our model is to actively learn with extra adversarial samples. We found this strategy particularly useful for further improving the feature robustness and discrminativity when incorporating with the sphere projection operation. In this work, we construct the adversarial sample $\vecx'$ for each  training sample $\vecx$ by the fast gradient sign method \cite{goodfellow_explaining_2014}:
\begin{equation}
        \vecx' = \vecx + \epsilon \cdot \frac{\nabla_{\vecx}\mathcal{L}'}{\| \nabla_{\vecx}\mathcal{L}'\|}
        \label{formula:data_aug}
\end{equation}
where $\nabla_{\vecx} \mathcal{L}'$ is the gradient of the learning objective w.r.t $\vecx$, $\epsilon$ is a small `permutation factor' taken as $\epsilon = 0.1$. Therefore $\vecx'$ is constructed such that it causes the value of the objective function to change the most quickly. In our method, the effect of learning with these adversarial samples can be understood as that they cause the hyperspheres of different classes to `oscillate' around their initial positions. As the center of these hyperspheres can only move on the surface of a hypersphere, this makes them easily crash with each other during optimization (Figure \ref{fig:3_2}.c). To avoid this, the optimizer would have to learn to stop such oscillations as well as increases the inter-class margin (Figure \ref{fig:3_2}.d). Note that the computation of $\nabla_{\vecx} \mathcal{L}'$ in \eqref{formula:data_aug} can be done in a similar way as in  $\nabla_{\W} \mathcal{L}'$.

\begin{algorithm}[H]
\caption{DVSP Training}
\label{alg:algorithm}
\textbf{Input}: Biometric data pairs $\{(\vecx_i, y_i)\}^n_{i=1}$ \\
\textbf{Output}: Network weights $\mathbf{W} = \{ \mathbf{W}_f, \mathbf{W}_g \}$ \\
\textbf{Hyperparamters}: IB factor $\beta$, projection radius $R$ \\
\begin{algorithmic}[1] 
\STATE sample $\{ \vecu_l \}^m_{l=1} \sim \mathcal{U}_{\text{sphere}}(\mathbf{0}, 1)$ ;
\WHILE{not converge}
        \FOR{$i = 1$ to $n$}
                \STATE construct adversarial sample $\vecx'_i$ from $\vecx_i$ by \eqref{formula:data_aug};
        \ENDFOR
        \STATE compute $\nabla_{\mathbf{W}}\mathcal{L}'$ with $\{(\vecx_i, \vecx'_i, y_i)\}^{n}_{i=1}$ and $\{ \vecu_l \}^m_{l=1}$ \;
        \STATE update weights: $\mathbf{W} \leftarrow$  Adam($\mathbf{W}$, $\nabla_{\mathbf{W}}\mathcal{L}'$) \;
\ENDWHILE
\STATE \textbf{return} $\W$
\end{algorithmic}
\end{algorithm}

The whole training procedure of the proposed deep variational sphere projection (DVSP) network is summarized in Algorithm 1. An overview of the full pipeline of the proposed feature extraction framework is given in Figure \ref{figure:nn-architecture}, where we will first use pre-processing and CNN to extract preliminary features before subsequent learning. After training, we take $\vecz'$ in \eqref{formula:z'} as the feature of $\vecx$, and compare two biometrics by the cosine distance between their features.

%% file: sections/4-experiments.tex
\begin{figure*}[t]
    \centering
        \includegraphics[width=1.00\textwidth]{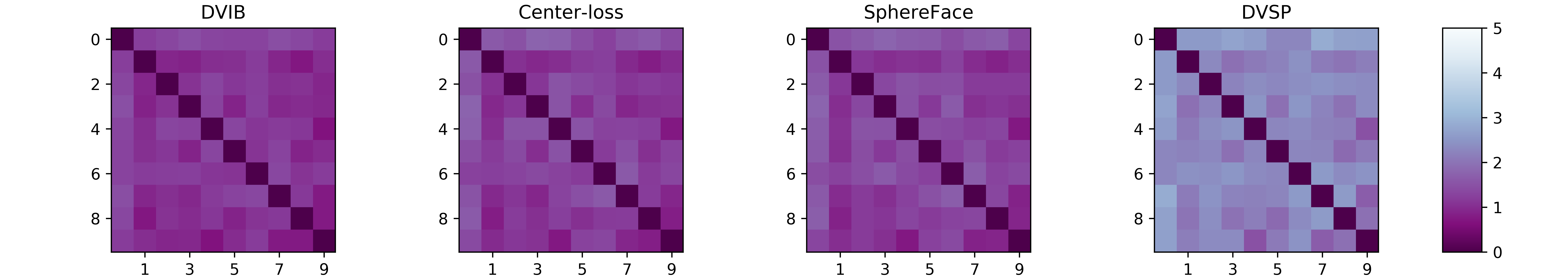}
        \caption{The robustness heatmap for casting adversarial attacks in each method for the MNIST dataset.}
        \label{figure:mnist-heatmap}
\end{figure*}

\section{Experiments}
This section setups a set of experiments to ascertain the effectiveness of the proposed DVSP method.

\subsection{Setup}

\noindent \textbf{Evaluation metric}. We will be using the following metrics for performance evaluation:
    
\begin{itemize}[leftmargin=*]
        \item \emph{Recognition performance}. We assess this by the equal error rate (EER) of the features. EER is the point where 
        the false positive rate (FPR) and the false negative rate (FNR) are equal. The FPR and FNR values are calculated as:
        \begin{equation}
            FPR = \frac{n_{FP}}{n_{FP} + n_{TN}}, \qquad  FNR = \frac{n_{FN}}{n_{FN} + n_{TP}}
        \end{equation}
        where $n_{TP}$, $n_{FP}$, $n_{TN}$, $n_{FN}$ are the number of true positive match, false positive match, true negative match and false negative match in the  database respectively. A lower EER indicates a better FPR under the same FNR and hence a better recognition performance.
        \item \emph{Feature robustness}. This metric measures how robust the feature extractor is against adversarial attacks. We compute this metric by averaging the robustness index $\mathcal{H}_{k,l}$ in \eqref{formula:pairwise-robustness-index} over all pairs of classes:
        \begin{equation}
    \begin{split}
                RI = \frac{2}{n_c(n_c-1)}  \sum_{l \neq k}^{n_c} \sum_{k=1}^{n_c} \mathcal{H}_{k,l} 
            \end{split}
    \label{formula:robustness-index}
\end{equation}
    where $n_c$ is the number of different individuals (classes) in the database. For example, if there are 10 individuals, we will calculate $\frac{10 \times 9}{2} = 45$ pairs of $\mathcal{H}_{k,l}$ values in \eqref{formula:robustness-index}. 
\end{itemize}

\noindent \textbf{Baselines}. We will compare the proposed method with the following approaches:

\begin{itemize}[leftmargin=*]
        \item \emph{Deep Variational Information Bottleneck} (DVIB) \cite{alemi_deep_2017} is a representation learning framework that extracts robust, highly discriminative features via the IB principle. In this work we adapt DVIB from classification tasks to the context of biometric recognition and as such it serves as the predecessor of our method.
        \item \emph{Center-loss} \cite{wen_discriminative_2016} is a powerful method for face recognition. It learns large-margin features by making points in each class to collapse towards the class center. 
        \item \emph{SphereFace} \cite{liu_sphereface:_2017} is another state-of-the-art method for face recognition. Similar to our method it projects the features onto the surface of a hypersphere. The difference is that it learns features by the softmax loss whereas ours is by the IB principle. 
\end{itemize}

All baselines as well as the proposed method adopt the same CNN as in \cite{liu_sphereface:_2017}, which is comprised of 20 layers. The IB factor $\beta$ as well as the projection radius $R$ in the proposed method are selected by 5-fold cross-validation such that it achieves the best balance between EER and RI. All neural networks are trained by Adam \cite{kingma_adam:_2014} with its default settings. 

\subsection{Results}

\noindent \textbf{Experiments on MNIST dataset}. We use this relatively simple dataset as a conceptual demonstration of the problem considered in this work. The MNIST dataset contains 60,000 images of hand-written digits from 0 to 9. In this experiment, we treat every digit as an individual in the biometric recognition system. Therefore there are 10 different individuals with 6,000 images for each. We use the first 50,000 samples for training and the left 10,000 for testing. 

Figure \ref{figure:mnist-heatmap} visualizes the 'robustness heatmap' of each method in this dataset. Each entry in the heatmap represents how hard it is to cast adversarial attacks between two classes i.e the value of $\mathcal{H}_{k,l}$. A brighter color of the entry indicates a larger $\mathcal{H}_{k,l}$ value and hence a higher difficulty. For example, the entry at the $4$-th row and the $6$-th column indicates the difficulty to attack between digit 3 and digit 5. Note that the average of the non-diagonal entries in the heatmap makes up the robustness index RI in \eqref{formula:robustness-index}. It can be seen from the heatmap that the proposed DVSP method arguably achieves better robustness than any other methods among almost all classes pairs. Most $\mathcal{H}_{k,l}$ values in the proposed DVSP method are above 2 but are below 1 in other methods. This indicates that it is nearly twice difficult to cast adversarial attacks in our method. We further discover that such difficulties indeed depend on the nature of the two classes. For example, the values of $\mathcal{H}_{4,9}$ and $\mathcal{H}_{3,5}$ are significantly much lower than other pairs. This coincides with the prior that it requires less efforts to cast adversarial attacks between similar classes.

\input{tables/table-mnist.tex}

Table \ref{table:MADR-mnist} further summarizes the EER and RI values of each method, from which we clearly see that the proposed DVSP method outperforms the other ones in both terms of recognition performance and feature robustness. 

\begin{figure*}[t]
    \centering
        \includegraphics[width=0.97\textwidth]{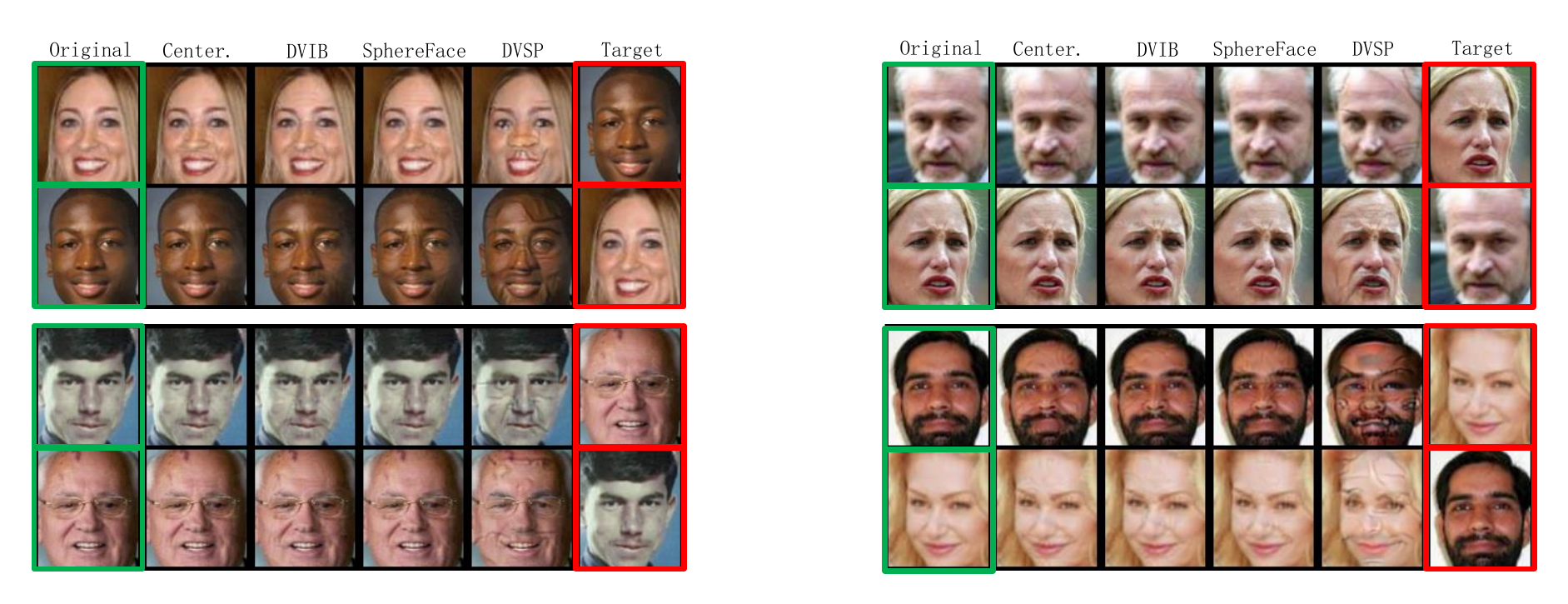}
        \caption{Visualizing the adversarial samples constructed by adversarial attacks in each method for the CASIA/LFW dataset. It can be seen that the adversarial samples in DVSP (the 5th columns) are significantly different from the original images (the 1st columns). }
        \label{figure:lfw}
\end{figure*}

\noindent \textbf{Experiments on CASIA/LFW dataset}.
In this part, we further evaluate the efficacy of the proposed method on a real-world face recognition problem. In particular, we consider the case that the individuals (classes) in the training set are disjoint from that in the test set; a common situation in real-world applications. Here, the CASIA-webface dataset \cite{yi_learning_2014} will serve as the training set and the LFW dataset \cite{huang2008labeled} will serve as the test set. There are 494,414 images of 10,575 individuals in  CASIA-webface and 13233 images of 5,749 individuals in the LFW dataset. Each of these images is pre-processed by the face alignment method \cite{zhang_joint_2016} before subsequent processing. 

In Figure \ref{figure:lfw}, we show four groups of images $\vecx_1, \vecx_2$ as well as their adversarial samples $\tilde{\vecx}_1, \tilde{\vecx}_2$ constructed by adversarial attacks. In each group, we display the original images in the first column, their corresponding adversarial samples in each method in the middle columns, and the target images in the last column. The first row in each pair shows the case $\vecx_1$ v.s $\vecx_2$ whereas the second row shows the case $\vecx_2$ v.s $\vecx_1$. These adversarial samples are found by the modified Carlini-Wanger attack described in section 2. It can be seen from the figure that the differences between the adversarial samples and their origin image are visually imperceptible in other approaches but are significant in the proposed DVSP method. The difference between the adversarial sample and the original image in DVSP is so large that it clearly deviates from natural human faces (e.g see the stripes on the faces in the 5-th columns in each group). This demonstrates that it requires much higher efforts to attack the proposed DVSP method than other methods i.e it is more robust to adversarial attacks.

\input{tables/table-LFW.tex}

Table \ref{table:MADR-lfw} further summarizes the EER and RI values of each method in this problem. Expectedly we see that the proposed DVSP achieves a better trade-off between recognition performance and feature robustness, with its RI index nearly three times higher than other methods and its EER much lower.

%% file: tables/table-mnist.tex

\begin{table}[h!]
  \centering
  \caption{The EER and the RI values on the MNIST dataset.}
    \begin{tabular}{lcccc}
    \toprule
         \quad \quad  & DVIB & Center-loss & SphereFace & DVSP \\
    \midrule
    EER \tiny{($\times 0.01$)} & 1.267  & 1.562 &  2.146 &  \textbf{0.725}  \\
    RI \quad  \quad & 1.111  & 1.245 &  1.299 &  \textbf{2.292}  \\
    \bottomrule
    \end{tabular}%
  \label{table:MADR-mnist}%
\end{table}%

%% file: tables/table-LFW.tex
\begin{table}[h!]
  \centering
  \caption{The EER and the RI values on the CASIA/LFW dataset.}
    \begin{tabular}{lcccc}
    \toprule
         \quad & DVIB & Center-loss & SphereFace & DVSP \\
    \midrule
    EER \tiny{($\times 0.01$)}  & 2.133  & 2.000 &  1.933 &  \textbf{1.593}  \\
    RI \quad & 1.393  & 0.612 &  0.572 &  \textbf{4.602}  \\
    \bottomrule
    \end{tabular}%
  \label{table:MADR-lfw}%
\end{table}%

%% file: sections/5-conclusion.tex
\vspace{0.05cm}

\section{Conclusions}
Adversarial attacks is a hot research topic in recent machine learning but has not been extensively studied within the context of biometric recognition. In this work, we extend current analysis in adversarial attacks from label space to feature space and discuss its effects on biometric recognition tasks. We discover that many state-of-the-art solutions in biometric recognition are indeed vulnerable to adversarial attacks and can be easily fooled. To defend such attacks, we develop a deep, information-theoretic framework to extract robust yet discriminative biometric features. The framework builds upon the recent deep variational information bottleneck method \cite{alemi_deep_2017} but is carefully adapted to biometric recognition tasks. Experiments on a toy dataset as well as a real-world face recognition problem demonstrates the efficacy of the proposed framework. To the best of our knowledge, only  few works like \cite{duan2018deep,xu2018deep_adversarial_metric} have considered adversarial attacks in the feature space but they have not studied the case of biometric recognition. Our work thus broadens the research scope of adversarial attacks and biometric recognition and suggests new research directions in both these fields.  

Throughout our work, we mainly focus on the $L_2$ attack case where we measure the similarity between the adversarial sample and the original image by their Euclidean distance. It is therefore of great interest to generalize our analysis to other  cases such as $L_0$ attack \cite{papernot_distillation_2016} and $L_{\infty}$ attack \cite{warde201611} for completeness. Another direction worth exploration is to test the effectiveness of the proposed framework on other biometric modalities not dealt with herein, including irises, voices and fingerprints.


%% file: ijcai19.bbl
\begin{thebibliography}{}

\bibitem[\protect\citeauthoryear{Alemi \bgroup \em et al.\egroup
  }{2017}]{alemi_deep_2017}
Alexander~A. Alemi, Ian Fischer, Joshua~V. Dillon, and Kevin Murphy.
\newblock Deep variational information bottleneck.
\newblock In {\em International Conference on Learning Representations}, 2017.

\bibitem[\protect\citeauthoryear{Amos \bgroup \em et al.\egroup
  }{2016}]{amos_openface_2016}
Brandon Amos, Bartosz Ludwiczuk, Mahadev Satyanarayanan, et~al.
\newblock Openface: A general-purpose face recognition library with mobile
  applications.
\newblock {\em CMU School of Computer Science}, 2016.

\bibitem[\protect\citeauthoryear{Bastani \bgroup \em et al.\egroup
  }{2016}]{bastani_measuring_2016}
Osbert Bastani, Yani Ioannou, Leonidas Lampropoulos, Dimitrios Vytiniotis,
  Aditya Nori, and Antonio Criminisi.
\newblock Measuring neural net robustness with constraints.
\newblock In {\em Advances in neural information processing systems}, pages
  2613--2621, 2016.

\bibitem[\protect\citeauthoryear{Carlini and
  Wagner}{2017}]{carlini_towards_2017}
Nicholas Carlini and David Wagner.
\newblock Towards evaluating the robustness of neural networks.
\newblock In {\em Security and {Privacy} ({SP}), 2017 {IEEE} {Symposium} on},
  pages 39--57. IEEE, 2017.

\bibitem[\protect\citeauthoryear{Chopra \bgroup \em et al.\egroup
  }{2005}]{chopra_learning_2005}
Sumit Chopra, Raia Hadsell, and Yann LeCun.
\newblock Learning a similarity metric discriminatively, with application to
  face verification.
\newblock In {\em Computer {Vision} and {Pattern} {Recognition}, 2005. {CVPR}
  2005. {IEEE} {Computer} {Society} {Conference} on}, volume~1, pages 539--546.
  IEEE, 2005.

\bibitem[\protect\citeauthoryear{Duan \bgroup \em et al.\egroup
  }{2018}]{duan2018deep}
Yueqi Duan, Wenzhao Zheng, Xudong Lin, Jiwen Lu, and Jie Zhou.
\newblock Deep adversarial metric learning.
\newblock In {\em Proceedings of the IEEE Conference on Computer Vision and
  Pattern Recognition}, pages 2780--2789, 2018.

\bibitem[\protect\citeauthoryear{Goodfellow \bgroup \em et al.\egroup
  }{2014}]{goodfellow_explaining_2014}
Ian~J. Goodfellow, Jonathon Shlens, and Christian Szegedy.
\newblock Explaining and harnessing adversarial examples.
\newblock {\em arXiv preprint arXiv:1412.6572}, 2014.

\bibitem[\protect\citeauthoryear{Huang \bgroup \em et al.\egroup
  }{2008}]{huang2008labeled}
Gary~B Huang, Marwan Mattar, Tamara Berg, and Eric Learned-Miller.
\newblock Labeled faces in the wild: A database forstudying face recognition in
  unconstrained environments.
\newblock In {\em Workshop on faces in'Real-Life'Images: detection, alignment,
  and recognition}, 2008.

\bibitem[\protect\citeauthoryear{Huang \bgroup \em et al.\egroup
  }{2017}]{huang_safety_2017}
Xiaowei Huang, Marta Kwiatkowska, Sen Wang, and Min Wu.
\newblock Safety verification of deep neural networks.
\newblock In {\em International {Conference} on {Computer} {Aided}
  {Verification}}, pages 3--29. Springer, 2017.

\bibitem[\protect\citeauthoryear{Kingma and Ba}{2014}]{kingma_adam:_2014}
Diederik~P. Kingma and Jimmy Ba.
\newblock Adam: {A} method for stochastic optimization.
\newblock {\em arXiv preprint arXiv:1412.6980}, 2014.

\bibitem[\protect\citeauthoryear{Kingma and
  Welling}{2013}]{kingma_auto-encoding_2013}
Diederik~P. Kingma and Max Welling.
\newblock Auto-encoding variational bayes.
\newblock {\em arXiv preprint arXiv:1312.6114}, 2013.

\bibitem[\protect\citeauthoryear{Liu \bgroup \em et al.\egroup
  }{2016}]{liu2016large}
Weiyang Liu, Yandong Wen, Zhiding Yu, and Meng Yang.
\newblock Large-margin softmax loss for convolutional neural networks.
\newblock In {\em ICML}, pages 507--516, 2016.

\bibitem[\protect\citeauthoryear{Liu \bgroup \em et al.\egroup
  }{2017}]{liu_sphereface:_2017}
Weiyang Liu, Yandong Wen, Zhiding Yu, Ming Li, Bhiksha Raj, and Le~Song.
\newblock Sphereface: {Deep} hypersphere embedding for face recognition.
\newblock In {\em The {IEEE} {Conference} on {Computer} {Vision} and {Pattern}
  {Recognition}}, volume~1, 2017.

\bibitem[\protect\citeauthoryear{Moosavi-Dezfooli \bgroup \em et al.\egroup
  }{2016}]{moosavi_deepfool_2016}
Seyed-Mohsen Moosavi-Dezfooli, Alhussein Fawzi, and Pascal Frossard.
\newblock Deepfool: a simple and accurate method to fool deep neural networks.
\newblock In {\em Proceedings of the IEEE Conference on Computer Vision and
  Pattern Recognition}, pages 2574--2582, 2016.

\bibitem[\protect\citeauthoryear{Papernot \bgroup \em et al.\egroup
  }{2016}]{papernot_distillation_2016}
Nicolas Papernot, Patrick McDaniel, Xi~Wu, Somesh Jha, and Ananthram Swami.
\newblock Distillation as a defense to adversarial perturbations against deep
  neural networks.
\newblock In {\em Security and {Privacy} ({SP}), 2016 {IEEE} {Symposium} on},
  pages 582--597. IEEE, 2016.

\bibitem[\protect\citeauthoryear{Schroff \bgroup \em et al.\egroup
  }{2015}]{schroff_facenet_2015}
Florian Schroff, Dmitry Kalenichenko, and James Philbin.
\newblock Facenet: {A} unified embedding for face recognition and clustering.
\newblock In {\em Proceedings of the {IEEE} conference on computer vision and
  pattern recognition}, pages 815--823, 2015.

\bibitem[\protect\citeauthoryear{Sun \bgroup \em et al.\egroup
  }{2014}]{sun_deep_2014}
Yi~Sun, Yuheng Chen, Xiaogang Wang, and Xiaoou Tang.
\newblock Deep learning face representation by joint
  identification-verification.
\newblock In {\em Advances in neural information processing systems}, pages
  1988--1996, 2014.

\bibitem[\protect\citeauthoryear{Sun \bgroup \em et al.\egroup
  }{2015}]{sun_deepid3_2015}
Yi~Sun, Ding Liang, Xiaogang Wang, and Xiaoou Tang.
\newblock Deepid3: {Face} recognition with very deep neural networks.
\newblock {\em arXiv preprint arXiv:1502.00873}, 2015.

\bibitem[\protect\citeauthoryear{Szegedy \bgroup \em et al.\egroup
  }{2013}]{szegedy_intriguing_2013}
Christian Szegedy, Wojciech Zaremba, Ilya Sutskever, Joan Bruna, Dumitru Erhan,
  Ian Goodfellow, and Rob Fergus.
\newblock Intriguing properties of neural networks.
\newblock {\em arXiv preprint arXiv:1312.6199}, 2013.

\bibitem[\protect\citeauthoryear{Tishby \bgroup \em et al.\egroup
  }{2000}]{tishby2000information}
Naftali Tishby, Fernando~C Pereira, and William Bialek.
\newblock The information bottleneck method.
\newblock {\em arXiv preprint physics/0004057}, 2000.

\bibitem[\protect\citeauthoryear{Warde-Farley and
  Goodfellow}{2016}]{warde201611}
David Warde-Farley and Ian Goodfellow.
\newblock 11 adversarial perturbations of deep neural networks.
\newblock {\em Perturbations, Optimization, and Statistics}, page 311, 2016.

\bibitem[\protect\citeauthoryear{Wen \bgroup \em et al.\egroup
  }{2016}]{wen_discriminative_2016}
Yandong Wen, Kaipeng Zhang, Zhifeng Li, and Yu~Qiao.
\newblock A discriminative feature learning approach for deep face recognition.
\newblock In {\em European {Conference} on {Computer} {Vision}}, pages
  499--515. Springer, 2016.

\bibitem[\protect\citeauthoryear{Xu \bgroup \em et al.\egroup
  }{2018}]{xu2018deep_adversarial_metric}
Xing Xu, Li~He, Huimin Lu, Lianli Gao, and Yanli Ji.
\newblock Deep adversarial metric learning for cross-modal retrieval.
\newblock {\em World Wide Web}, pages 1--16, 2018.

\bibitem[\protect\citeauthoryear{Yi \bgroup \em et al.\egroup
  }{2014}]{yi_learning_2014}
Dong Yi, Zhen Lei, Shengcai Liao, and Stan~Z. Li.
\newblock Learning face representation from scratch.
\newblock {\em arXiv preprint arXiv:1411.7923}, 2014.

\bibitem[\protect\citeauthoryear{Yuan \bgroup \em et al.\egroup
  }{2019}]{yuan2019adversarial}
Xiaoyong Yuan, Pan He, Qile Zhu, and Xiaolin Li.
\newblock Adversarial examples: Attacks and defenses for deep learning.
\newblock {\em IEEE transactions on neural networks and learning systems},
  2019.

\bibitem[\protect\citeauthoryear{Zhang \bgroup \em et al.\egroup
  }{2016}]{zhang_joint_2016}
Kaipeng Zhang, Zhanpeng Zhang, Zhifeng Li, and Yu~Qiao.
\newblock Joint face detection and alignment using multitask cascaded
  convolutional networks.
\newblock {\em IEEE Signal Processing Letters}, 23(10):1499--1503, 2016.

\end{thebibliography}
